\documentclass[letterpaper,10pt,journal,twoside]{IEEEtran}

\usepackage{amsmath,amssymb,amsfonts}
\usepackage{graphicx}
\usepackage{url}
\usepackage{stfloats}
\usepackage{booktabs}
\usepackage{siunitx}
\usepackage{cite}
\usepackage{hyperref}

\newcommand{\magflow}{\textit{Magflow}}
\newcommand{\piNC}{\ensuremath{\pi_{\mathrm{BC}}}}
\newcommand{\piPHY}{\ensuremath{\pi_{\mathrm{MF}}}}
\newcommand{\piPC}{\ensuremath{\pi_{\mathrm{PC}}}}
\allowdisplaybreaks

\newcommand{\ieeenotice}{%
  \parbox{0.86\paperwidth}{\centering\scriptsize
  This article has been accepted for publication in IEEE Robotics and Automation Letters.
  \copyright\,2026 IEEE. Personal use of this material is permitted. Permission from IEEE must be obtained for all other uses, in any current or future media, including reprinting/republishing this material for advertising or promotional purposes, creating new collective works, for resale or redistribution to servers or lists, or reuse of any copyrighted component of this work in other works.}}
\AddToHookNext{shipout/foreground}{%
  \put(0.5\paperwidth,-\dimexpr\paperheight-0.28in\relax){\makebox[0pt][c]{\ieeenotice}}}

\begin{document}

\setlength{\abovedisplayskip}{4pt}\setlength{\belowdisplayskip}{4pt}%
\setlength{\abovedisplayshortskip}{0pt}\setlength{\belowdisplayshortskip}{0pt}%

\bstctlcite{IEEEexample:BSTcontrol}

\title{Magflow: Differentiable Simulation of\\Industrial-Grade 6D Magnetic Levitation}
\author{Philipp Hartmann, Jannick Stranghöner, and Klaus Neumann%
\thanks{Manuscript received: July 16, 2026; Revised October 7, 2026; Accepted November 4, 2026.\par\noindent\hspace*{1em}
This paper was recommended for publication by Editor Sehoon Ha upon evaluation of the Associate Editor and Reviewers' comments. This work was supported by the Fraunhofer Internal Programs under Grant No. SME 40-09551. (Corresponding author: Philipp Hartmann.)\par\noindent\hspace*{1em}
Philipp Hartmann, Jannick Stranghöner, and Klaus Neumann are with CITEC, Faculty of Technology, Bielefeld University, 33615 Bielefeld, Germany (e-mail: philipp.hartmann@uni-bielefeld.de). Klaus Neumann is also with Fraunhofer IOSB-INA, 32657 Lemgo, Germany.}%

}

\markboth{IEEE Robotics and Automation Letters. Preprint Version. Accepted November, 2026}%
{Hartmann \MakeLowercase{\textit{et al.}}: Magflow: Differentiable Simulation of Industrial-Grade 6D Magnetic Levitation}

\maketitle

\begin{abstract}

Magnetic levitation with six-degrees-of-freedom enables frictionless transport and micrometer-scale positioning in principle, but low-level control is inherently challenging due to its complex and unstable dynamics. Industrial systems typically rely on deep hardware--software co-design, where mover and stator geometries are jointly designed to enable a compact, hand-crafted approximation of the electromagnetic force computation, thereby keeping control computationally tractable at kilohertz rates. However, this approach imposes overly conservative constraints on the operating workspace and the range of supported mover topologies. Instead of reducing the underlying physics to a hand-crafted approximation, we compute the physics directly. We introduce \magflow{}, a fully differentiable, GPU-parallel simulator for 6D magnetic levitation. It is used to train the first neural controller for an industrial-grade 6D magnetic levitation system that is learned purely from simulated physics. We show that the learned neural controller transfers zero-shot from simulation to real hardware while satisfying hard real-time constraints at 4 kHz. The learned controller achieves qualitatively new behaviors and even levitates a mover topology for which no hand-crafted solution exists. These results demonstrate that co-design is not a physical necessity for 6D magnetic levitation control. More broadly, control learning from simulated physics can find solutions beyond the reach of control engineering in demanding industrial domains. Source code and demonstration videos are publicly available at \mbox{\texttt{\url{https://sites.google.com/view/magflow}}}.

\end{abstract}


\section{Introduction}
\label{sec:intro}

\IEEEPARstart{M}{agnetic} levitation (MagLev) with six-degrees-of-freedom (6D) enables frictionless transport, precise manipulation~\cite{BergmannTransportation2026}, and flexible product routing with no mechanical wear. These properties make MagLev systems attractive for semiconductor manufacturing, pharmaceutical filling, electronics assembly, and many other industrial applications~\cite{LuUsman2012,ZhouWu2022,BeckhoffXPlanar2026}. Today, these systems achieve micrometer-scale positioning~\cite{BeckhoffAPS4322}, but this precision took decades of research.

\noindent
In such systems, a mover with embedded permanent magnets levitates above planar stator tiles. The controller estimates its pose from magnetic field sensors and computes the currents of their coils to stabilize the inherently unstable mover in 6D. The reason these systems work is their deep hardware-software co-design. In industrial systems, mover and stator geometries are typically jointly designed so that the forces between coils and magnets admit compact analytical expressions~\cite{LuUsman2012,Kim1998,Jansen2008,VanLierop2008}, enabling hard real-time control at kilohertz rates~\cite{deBoeijLomonova2009,deBoeij2006}. Alternatively, the forces can be precomputed numerically for a given geometry and interpolated online~\cite{deBoeijLomonova2009,BerkelmanDzadovsky2013,BerkelmanLu2020}. Either way, the resulting pipeline is a chain of engineered modules, from pose estimation and commutation to feedback control and calibration~\cite{Kim1998,Jansen2008,VanLierop2008,XuLu2020}, each built on geometry-specific assumptions and carefully tuned to reduce error compounding across the control chain.

\begin{figure}[t]
  \centering
    \includegraphics[width=\linewidth]{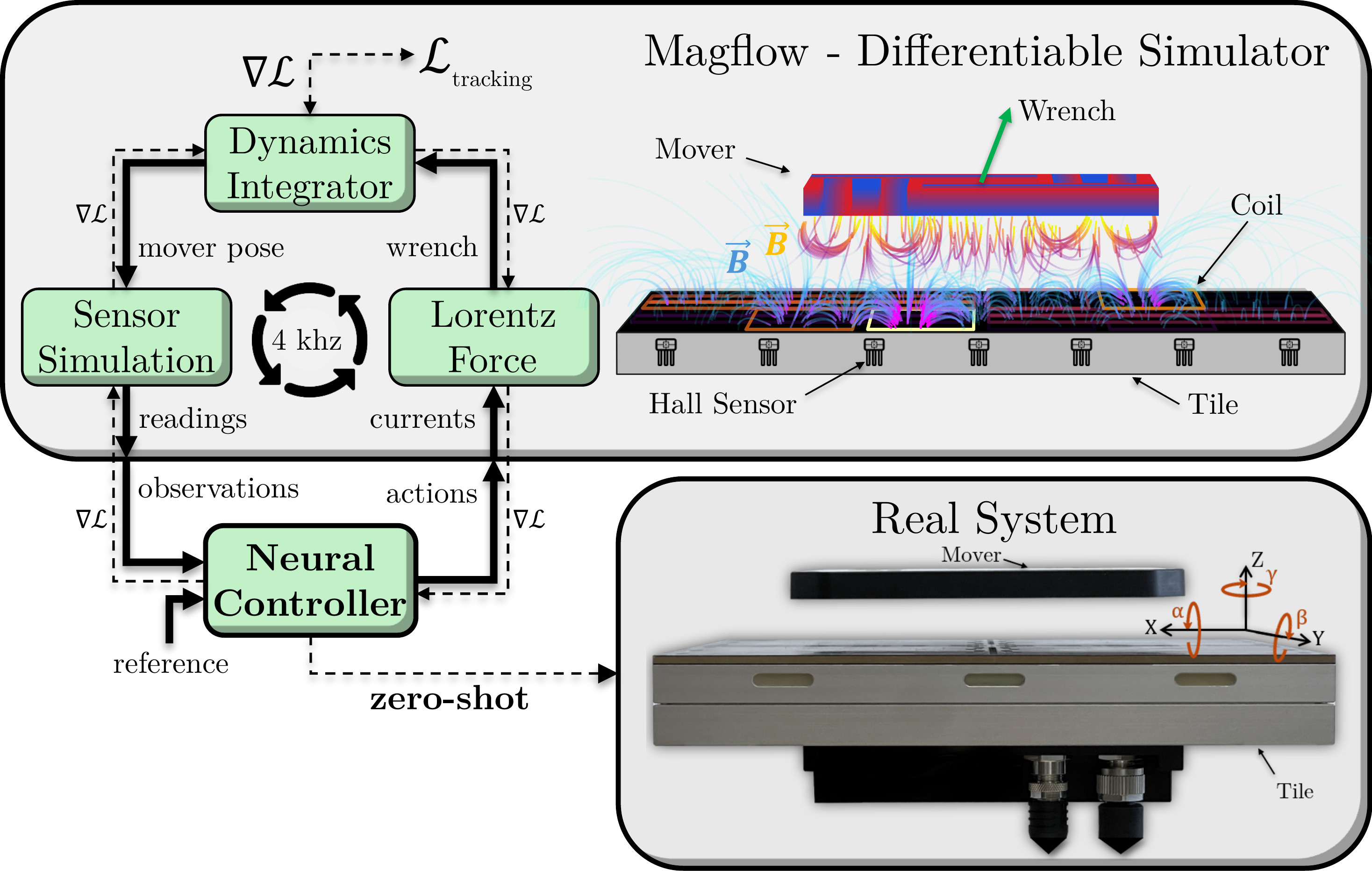}
    \vspace{-2.0em}
    \caption{%
      \magflow{} is the first differentiable simulator for 6D magnetic levitation, fusing the electromagnetic physics and the neural controller into a single GPU computational graph. Backpropagation through time trains the controller end-to-end, purely from physics. The trained controller then transfers zero-shot to the real hardware running at \SI{4}{\kilo\hertz}.%
    }
    \vspace{-1.6em}
    \label{fig:VA}
\end{figure}

\noindent
Our approach is to move approximations out of the real-time control loop and into offline training, where its cost no longer matters. To this end, we introduce \magflow{} (see Fig.~\ref{fig:VA}), a fully differentiable, GPU-parallel electromagnetic simulator that models the Lorentz interactions between stator coils and mover magnets, the dynamics of the mover, and the Hall sensor measurements. Implemented entirely in JAX, \magflow{} compiles the physics, the neural controller, and the full closed-loop rollout into a single GPU computational graph, enabling backpropagation through time across parallel simulated worlds. Using \magflow{}, we train the first neural controller for 6D MagLev learned purely from simulated physics and transfer it zero-shot to the real hardware at \SI{4}{\kilo\hertz}. Because \magflow{} does not rely on hand-crafted approximations, the learned neural controller is not bound by their validity region and extends the yaw range of the industrial system beyond its engineered limit. In addition, we levitate a plain bar magnet, which does not meet the co-design assumptions and for which no engineered controller exists on this platform. While single disc magnets are limited to 5D control by their rotational symmetry~\cite{BerkelmanDzadovsky2010Novel,Berkelman2026}, the bar magnet is controlled in full 6D, even through a continuous \ang{360} rotation about a horizontal axis, which to our knowledge has not been shown for any single permanent magnet on a planar MagLev system.

\section{Related Work}
\label{sec:related}

\paragraph{Low-level control of 6D MagLev}
\label{sec:relatedMagLev}
One of the main challenges in 6D MagLev is to compute the currents which apply a desired wrench to the mover. The idea is to reduce the Lorentz-force integral to a compact model. For a 1D Halbach array, the force decouples into vertical lift and lateral thrust~\cite{Kim1998}. This model was extended to full 6D wrenches for 2D orthogonal Halbach layouts~\cite{Jansen2008}, to model-based commutation for long-stroke actuators~\cite{VanLierop2008}, and to multi-meter travel with linearly scaling coil counts~\cite{LuUsman2012}. Later work refined this model with yaw-dependent wrenches, unified wrench models, and predictor-based control~\cite{XuLu2020,Zhu2017,UWM2018,EUWM6DOF2024,PredictorMPC2023}. Others reduced the remaining model errors with compensation models~\cite{UsmanLu2015,DataDrivenFT2022} or by redesigning the magnet topology with gap or staggered Halbach arrays~\cite{Zhu2015ForceHarmonics,Wang2018GapsStaggers}. Such compensation can substantially improve accuracy, for instance by roughly \(50\%\) in force and \(70\%\) in torque over a harmonic model~\cite{Zou2023ForceTorque}.

\begin{figure*}[t]
  \centering
  \includegraphics[width=\textwidth]{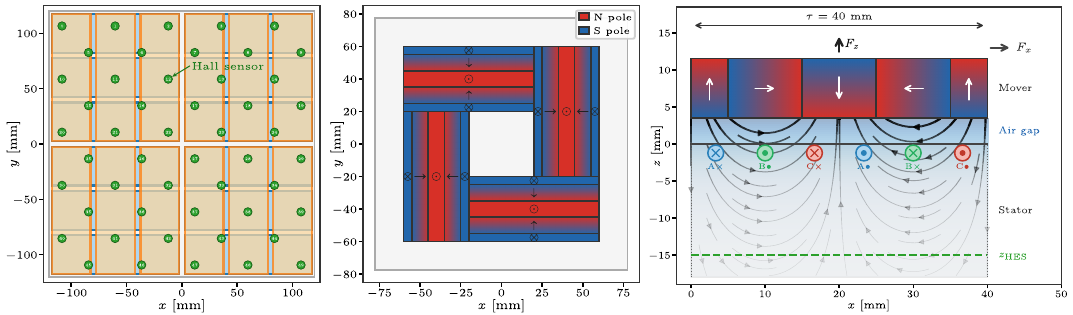}
  \vspace{-2.0em}
\caption{%
  Overview of the MagLev system~\cite{BeckhoffPatentStator2022,BeckhoffPatentSensor2022} used in this work.
  \textbf{(left)}~APS4322 stator tile with 24 stator units arranged in horizontal and vertical layers and the 50 embedded triaxial HES.
  \textbf{(middle)}~Bottom view of the APM4330 mover with four five-magnet Halbach arrays, where red and blue indicate north and south pole.
  \textbf{(right)}~Cross-section of a vertical stator unit beneath an aligned vertical Halbach array, showing the three-phase conductors, the HES plane, and the resulting magnetic flux. \(F_z\) and \(F_x\) indicate lift and lateral thrust.%
}
  \vspace{-1.6em}
  \label{fig:system_overview}
\end{figure*}

\noindent
Alternatively, forces can be precomputed offline and stored in lookup tables, whose size grows exponentially with the pose dimensions, so they are kept tractable by exploiting the symmetry of the magnet array~\cite{deBoeijLomonova2009}. Research systems levitate disc magnets above cylindrical coils, whose radial symmetry reduces such tables to a single magnet--coil pair, using optical tracking and PD control~\cite{BerkelmanDzadovsky2010Novel,BerkelmanDzadovsky2013,BerkelmanLu2020,Berkelman2026}. A single disc magnet receives no torque about its symmetry axis, limiting it to 5D~\cite{BerkelmanDzadovsky2010Novel,Berkelman2026}, and even a vertical orientation is only marginally feasible without overheating the coils~\cite{BerkelmanDzadovsky2010Novel}. Platforms of two discs control all 6D, but coil heating limits their tilt to \(\pm\ang{40}\)~\cite{BerkelmanDzadovsky2013}, while unlimited rotation required a body of six magnets, arranged such that the required coil currents remain bounded in every orientation, levitated at a fixed height with optical tracking~\cite{MiyasakaBerkelman2014}. All share one structure: an engineered actuation model, made tractable by co-design or symmetry, with software tuned and calibrated around it. In contrast, we rotate a single bar magnet, sensed only by the Hall sensors embedded in an industrial MagLev system. As neither the bar nor the planar rectangular coils have a continuous rotational symmetry, such a table needs more pose dimensions and becomes substantially larger, possibly intractable for real-time control~\cite{deBoeijLomonova2009}.

\noindent
Recent research replaced an engineered controller with a single end-to-end neural controller learned via behavior-cloning from interaction data of the proprietary controller~\cite{Hartmann2025}. This proved feasibility, but only because a working engineered controller already existed to imitate. The neural controller could not surpass its teacher and inherited the conservative workspace of the engineered system. We instead learn control directly from physics, with no commutation model and no expert to imitate.

\paragraph{Neural control of unstable, high-rate systems}
Neural networks can learn control directly from data, performing well in domains that share MagLev's traits: open-loop instability, high actuation rates, and tight sensor--actuator coupling. Neural controllers have been trained for quadrotor flight~\cite{lambertLowLevelControlQuadrotor2019,feredeEndtoendNeuralNetwork2024} and for electromagnetic systems such as electric motors and tokamak plasma control~\cite{schenkeDeepQLearning2021,Degrave2022}.
They also outperform modular pipelines in manipulation, musculoskeletal control, and industrial assembly~\cite{ShareRL2025,levineEndtoEndTraining2016,driessLearningControl2018}.
However, none meets MagLev's full combination at once: six unstable degrees of freedom, 72 coupled actuator commands, 150 raw magnetic-field observations with no direct state signal, \SI{4}{\kilo\hertz} hard real-time control, and micrometer-scale precision.

\paragraph{Differentiable simulation and BPTT}
Differentiable simulation enables gradient-based optimization of system dynamics, allowing a controller to be trained via backpropagation through time (BPTT) across the rollout~\cite{Werbos1990,BelbutePeres2018}.
GPU-parallel differentiable simulators enable large-batch policy optimization for locomotion~\cite{Brax2021} and most prominently for agile flight~\cite{CrazyFlow2026,Heeg2025,PanLearningFly2026}, although stiffness or discontinuities may compromise the efficacy of first-order gradient estimates~\cite{Suh2022}.
Controllers trained this way already fly real quadrotors~\cite{CrazyFlow2026,PanLearningFly2026} and can outperform engineered adaptive control~\cite{PanLearningFly2026}. However, these works differentiate through rigid-body dynamics at a few hundred \si{\hertz}, whereas 6D MagLev requires differentiating through far more complex electromagnetic interactions between coils, mover magnets, and Hall sensors at kilohertz rates. \magflow{} is the first simulator to achieve this.

\paragraph{Sim-to-real transfer}
Domain randomization is the standard method to tackle the sim2real gap between simulation and hardware. Here, training across a distribution of physical parameters yields controllers that transfer zero-shot to the real world~\cite{Tobin2017}. It bridges simulation and reality in many domains such as dexterous in-hand manipulation~\cite{Andrychowicz2019}, locomotion across morphologies~\cite{Peng2018}, and end-to-end nano-quadcopter control~\cite{GraefeCFB2026}. In each case it covers the real system's unknown physical parameters with a distribution in simulation. Our system has many such uncertainty sources, discussed in Sec.~\ref{subsec:domain_rand}. Aligning with previous work, we also address these uncertainties using domain randomization.

\section{System Platform and Engineering}
\label{sec:system}

\subsection{Hardware Platform}
\label{sec:system_hardware}

\noindent
In this work, we consider a single-tile, single-mover configuration. The hardware platform is a Beckhoff XPlanar system comprising an active tile (APS4322-0000-0000, \(240 \times 240\,\si{\milli\metre}\)) and a passive mover (APM4330-0000-0000, \(155 \times 155\,\si{\milli\metre}\)) (see Fig.~\ref{fig:system_overview}). For implicit state feedback, an embedded array of 50 triaxial Hall-effect sensors (HES) provides 150 raw magnetic field values per control cycle~\cite{BeckhoffAPS4322,BeckhoffPatentSensor2022}. Depending on the mover pose, these readings empirically range from \(-2000\) to \(2000\) raw sensor counts (\(\pm\SI{0.1}{\tesla}\)) with an inherent noise standard deviation of approximately 4 least-significant bits.
For actuation, the tile contains 24 PCB stator units arranged as 12 horizontal and 12 vertical units across multiple copper layers~\cite{BeckhoffPatentStator2022}. Each unit consists of three phase coils whose active conductors follow the \(A^+B^-C^+A^-B^+C^-\) order~\cite{BeckhoffPatentStator2022} shown in Fig.~\ref{fig:system_overview} (right), and is driven via field-oriented control using a distinct control triplet of direct current \(d_j\), quadrature current \(q_j\), and electrical angle \(\phi_j\). To minimize bus payload, currents are quantized as 16-bit signed integers and electrical angles as 16-bit unsigned integers, yielding 72 actuator commands per control step for a single tile.

\noindent
The passive mover (see Fig.~\ref{fig:system_overview}) comprises four orthogonal five-magnet Halbach arrays, each spanning approximately \(80 \times 40\,\si{\milli\metre}\)~\cite{BeckhoffAPM4330,BeckhoffPatentSensor2022}. Full \ang{360} yaw rotation is only supported at specific symmetric positions centered on one, two, or four tiles~\cite{BeckhoffRotationLoadCapacity,BeckhoffXPlanar2026}. Outside these defined points, maximum yaw is generally restricted to \(\pm\ang{5}\)~\cite{BeckhoffAPS4322}. The XPlanar system is constrained to a control cycle of \SI{4}{\kilo\hertz}~\cite{BeckhoffTF5890}, executed within a single TwinCAT 3.1 real-time task, so any controller must compute its actuator commands within \SI{250}{\micro\second}.
To ensure deterministic execution times without hardware acceleration, all computation is restricted to one CPU core of an industrial PC (C6030, Intel i9-9900K) running in kernel mode independently of the host operating system.

\subsection{How Harmonic Commutation Enables Real-Time Control}

\noindent
MagLev controllers usually consist of manually engineered modules like pose estimation, feedback control, and commutation, all of which must be evaluated within a single \SI{250}{\micro\second} control cycle (Sec.~\ref{sec:system_hardware}). The pose estimation module infers the 6D mover pose \(\mathbf{p}\) from the HES readings~\cite{schultePlanarmotorantriebssystemXPlanar2022b}. The feedback controller then computes a 6D wrench \(\mathbf{w}^{\star}\), which must be realized as current commands for all stator units. The electromagnetic force of one magnet is obtained from the Lorentz integral over the stator conductor volume \(V_c\),
\begin{equation}
\mathbf{F}(\mathbf{p}) =
\int_{V_c} \mathbf{J}(\mathbf{r})\times\mathbf{B}(\mathbf{r};\mathbf{p})\,
\mathrm{d}^3\mathbf{r},
\label{eq:lorentz_integral}
\end{equation}
with point \(\mathbf{r}\) in \(V_c\), stator current density \(\mathbf{J}\), and mover field \(\mathbf{B}\). Evaluating this integral accurately online is computationally too expensive for real-time control~\cite{deBoeijLomonova2009}. To enable real-time control, industrial MagLev typically exploits hardware--software co-design. Mover and stator are designed such that this integral is well approximated by an analytical harmonic commutation model~\cite{Kim1998,Jansen2008,LuUsman2012,VanLierop2008}.

\noindent
The harmonic commutation model assumes that each stator unit interacts only with the Halbach array of matching orientation, so it is sufficient to consider one such pair. For an infinite and isolated Halbach array with ideal magnetic field~\cite{jinGeneralAnalyticalMethod2013}, it retains only the first harmonic of the field and of the stator current distribution and assumes aligned full-period coverage,
\begin{equation}
\begin{aligned}
B_x&=\hat B_1e^{-kz}\sin\eta,\qquad B_z=\hat B_1e^{-kz}\cos\eta,\\
J_y&=\hat J_1\,(d\sin\eta-q\cos\eta),
\end{aligned}
\label{eq:first_harmonic_fields}
\end{equation}
in stator coordinates \(x,y,z\) of a vertical stator unit with conductors along \(y\) and aligned array wave direction \(x\) (see Fig.~\ref{fig:system_overview}), with pole pitch \(\tau\), wavenumber \(k=2\pi/\tau\), air gap \(z\), array offset \(x_0\), electrical coordinate \(\eta=k(x-x_0)\), and amplitudes \(\hat B_1\), \(\hat J_1\). The rectangular magnets and discrete conductor segments introduce odd higher harmonics, which this model neglects. Combining these terms in the Lorentz integral~\eqref{eq:lorentz_integral} gives the levitation and thrust forces of the harmonic commutation
\begin{equation}
F_z(\mathbf{p}) = \alpha d,\qquad F_x(\mathbf{p}) = \alpha q,\qquad \alpha=\pi\hat B_1 e^{-kz}\hat J_1,
\label{eq:dq_decoupled}
\end{equation}
under which \(d\)-currents map purely to levitation forces and \(q\)-currents purely to lateral thrusts. The two vertical and two horizontal arrays of the mover thus yield local force pairs \((F_z,F_x)\) and \((F_z,F_y)\), stacked into \(\mathbf{f}\in\mathbb{R}^8\). Each pair acts at a fixed position relative to the mover center of mass, so a pose-independent matrix \(\mathbf{G}\) maps \(\mathbf{f}\) to the wrench \(\mathbf{w}^{\star}=\mathbf{G}\,\mathbf{f}\).

\noindent
A stator unit \(j\) overlapped by a Halbach array inherits its \((d_j,q_j)\) current command, while units with no overlap receive \((d_j,q_j)=(0,0)\). With the electrical angle \(\phi_j(\mathbf{p})=k\,(x_j-x_0)\) at stator unit position \(x_j\), the hardware internally computes the currents \(I_j^s\) of the three phase coils \(s\in\{A,B,C\}\) of each stator unit by
\begin{equation}
I_j^s = d_j\sin(\phi_j-\delta_s)+q_j\cos(\phi_j-\delta_s),
\label{eq:dq_currents}
\end{equation}
with coil offsets \(\delta_A=0\), \(\delta_B=-2\pi/3\), and \(\delta_C=2\pi/3\). This compact commutation step can be evaluated in hard real-time.

\subsection{Model Mismatch and Engineering Limits}
\label{subsec:model_mismatch}

\noindent
Control engineering exploits Eq.~\eqref{eq:dq_decoupled} by assuming first-harmonics and aligned full-period coverage. However, the true wrench deviates due to many neglected effects: higher harmonics, finite-array and end fields, neighboring-array coupling, nonzero across-field components, discrete-conductor layout, and manufacturing tolerances. Together, these errors appear as force ripple, gain errors, and pose-dependent cross-coupling. Thus, \(d\)-current also produces thrust and \(q\)-current lift.

\noindent
In addition, manually engineered controllers are highly geometry-specific. Co-design of stator units and Halbach arrays assumes that the wave direction of each Halbach array stays aligned with the conductor direction of its stator unit, which holds only at small yaw. For a mover rotated by yaw \(\psi\), with array offset \((x_0,y_0)\), the electrical coordinate \(\eta=k\big((x-x_0)\cos\psi+(y-y_0)\sin\psi\big)\) varies along each conductor, and the in-plane array field along \(x\) is scaled by \(\cos\psi\). Integrating along the conductor length under the array, about one period \(\tau\) for the APM4330, turns Eq.~\eqref{eq:dq_decoupled} into
\begin{equation}
F_z=\alpha d\cos\psi\,\mathrm{sinc}(\pi\sin\psi),\quad F_x=\alpha q\,\mathrm{sinc}(\pi\sin\psi),
\label{eq:yaw_gain}
\end{equation}
with \(\mathrm{sinc}(u)=\sin u/u\).
At \(\psi=\ang{5}\), the lift of the aligned units is still \SI{98}{\percent} of its nominal value, but at \ang{30} only \SI{55}{\percent}, and at \ang{45} only \SI{25}{\percent}. The commutation model thus allocates current less efficiently at larger yaw angles, and more current is needed for the same force. This is tractable, but it requires additional model structure, calibration, or compensation (Sec.~\ref{sec:related}). However, each such addition is another module whose own mismatch can contaminate the control pipeline. Therefore, the yaw angle is usually constrained (Sec.~\ref{sec:system_hardware}).

\noindent
A plain bar magnet does not meet these co-design assumptions, as it has no Halbach structure from which a harmonic commutation model could be derived, and it lacks the continuous rotational symmetry that keeps lookup tables tractable (Sec.~\ref{sec:related}). Therefore, no engineered controller exists for such a "mover" on this platform, which makes it a direct test of whether control can be achieved without an engineered actuation model. Both limitations, the constrained yaw range and the restriction to co-designed topologies, are therefore not directly physical constraints of the mover but rather consequences of the engineered actuation model that makes the software real-time capable.

\section{Magflow}
\label{sec:method}

\noindent
Data-driven learning removes the reliance on real-time-tractable approximations and hand-engineered models. Instead, a differentiable physics simulator computes the full electromagnetic physics offline, where real-time capability no longer matters, and provides gradients for the controller's actions.

\noindent
However, no such simulator exists for 6D MagLev. We therefore introduce \magflow{}, implemented entirely in JAX to exploit its automatic differentiation and just-in-time GPU compilation. This lets \magflow{} compile the full closed-loop rollout (physics and neural controller) into a single GPU computational graph, running many environments in parallel. In each rollout, given the controller's prediction, \magflow{} computes the full Lorentz force, integrates the mover dynamics, and samples the Hall sensors that form the controller's next input (see Fig.~\ref{fig:VA}). Differentiating through the rollout trains the neural controller by backpropagation through time (BPTT).

\subsection{Differentiable Electromagnetic Simulation}

\noindent
\textbf{Overview.} \magflow{} computes the field at the Hall sensors, observed by the controller, and the force and torque on the mover, which drive its dynamics. Both arise from two superimposable, differentiable field sources, the mover's permanent magnets and the stator's coils.

\noindent
\textbf{Frames.} Positions \(\mathbf{r}\) are expressed in a world frame with origin at the center of the tile surface and \(z\) pointing upward. The mover pose \(\mathbf{p}_t=(\mathbf{s}_t,\mathbf{R}_t)\) places the mover frame at the center of the mover's bottom face \(\mathbf{s}_t\) with orientation \(\mathbf{R}_t\in SO(3)\).

\noindent
\textbf{Mover field.} The mover holds \(M\) magnets \(m=1,\dots,M\) (\(M=20\) for the APM4330, \(M=1\) for the bar magnet). Each is a uniformly magnetized cuboid with center \(\boldsymbol{\rho}_m\), side lengths \(2a_m\times2b_m\times2c_m\) along the mover axes, and magnetization \(\mathbf{M}_m\), all in mover coordinates (remanence \(B_r\approx\SI{1.3626}{\tesla}\) for the APM4330, \(B_r\approx\SI{1.395}{\tesla}\) for the bar magnet). At a point \(\tilde{\mathbf{r}}\) relative to its center, a cuboid with half sizes \(a,b,c\) and magnetization \(\mathbf{M}\) generates the closed-form Akoun--Yonnet field~\cite{AkounYonnet1984}. With the corner signs \(\boldsymbol{\sigma}=(\sigma_x,\sigma_y,\sigma_z)\in\{-1,+1\}^3\) and the corner vectors \(\mathbf{u}_{\boldsymbol{\sigma}}=\tilde{\mathbf{r}}+\boldsymbol{\sigma}\odot(a,b,c)\), it is
\begin{equation}
  \mathbf{b}_{\mathrm{cub}}(\tilde{\mathbf{r}})
  = \frac{1}{4\pi} \sum_{\boldsymbol{\sigma}} \sigma_x \sigma_y \sigma_z\,
  \boldsymbol{\beta}(\mathbf{u}_{\boldsymbol{\sigma}};\mathbf{M})
  \label{eq:cuboid_corner_sum}
\end{equation}
where \(\boldsymbol{\beta}\) is the Akoun--Yonnet primitive. Writing \(\mathbf{b}_m\) for the field of Eq.~\eqref{eq:cuboid_corner_sum} with the parameters of magnet \(m\), transforming a world point into mover coordinates, and rotating the sum back gives the mover field \(\mathbf{B}^{\mathrm{mover}}_t(\mathbf{r})\in\mathbb{R}^3\),
\begin{equation}
  \mathbf{B}^{\mathrm{mover}}_t(\mathbf{r})
  = \mathbf{R}_t \sum_{m=1}^{M}
    \mathbf{b}_m\big(\mathbf{R}_t^\top(\mathbf{r}-\mathbf{s}_t)-\boldsymbol{\rho}_m\big).
  \label{eq:bfield_superposition}
\end{equation}

\noindent
\textbf{Stator field.} Each phase coil consists of rectangular conductor tracks stacked across multiple copper layers.
To resemble the coils closely, we represent the PCB coils by four layers and the width of each track by five parallel thin wires, each divided into straight segments of \SI{3}{\milli\meter}. Compared with 20 wires and \SI{1}{\milli\meter} segments, this changes the wrench by about \SI{1}{\percent}, and the number of layers matters more than the number of wires.
Segment \(\ell\) belongs to stator unit \(j\) and phase \(s\) and runs from \(\mathbf{r}_\ell^-\) to \(\mathbf{r}_\ell^+\) along its current \(I_j^s\) of Eq.~\eqref{eq:dq_currents}, with \(\Delta\mathbf{l}_\ell=\mathbf{r}_\ell^+-\mathbf{r}_\ell^-\). With \(\mathbf{d}^\pm=\mathbf{r}-\mathbf{r}_\ell^\pm\), the Biot--Savart law gives its field~\cite{Griffiths2017}
\begin{equation}
  \mathbf{b}_\ell(\mathbf{r})
  = \frac{\mu_0 I_j^s}{4\pi}\,
    \frac{\Delta\mathbf{l}_\ell\times\mathbf{d}^-}{\|\Delta\mathbf{l}_\ell\times\mathbf{d}^-\|^2}\,
    \Delta\mathbf{l}_\ell\cdot\left(\frac{\mathbf{d}^-}{\|\mathbf{d}^-\|}-\frac{\mathbf{d}^+}{\|\mathbf{d}^+\|}\right),
  \label{eq:biot_savart_segment}
\end{equation}
where \(\mu_0\) is the vacuum permeability. Since coils and sensors are fixed to the tile and the field is linear in the current, we precompute it once at the 50 triaxial HES as a sensitivity matrix \(\mathbf{S}\in\mathbb{R}^{150\times72}\). With the 72 phase currents \(I_j^s\) stacked into \(\mathbf{i}_t\in\mathbb{R}^{72}\), the stator field at the HES is
\begin{equation}
  \mathbf{B}^{\mathrm{stator}}_t(\mathbf{i}_t) = \mathbf{S}\,\mathbf{i}_t.
  \label{eq:stator_field_sensitivity}
\end{equation}

\noindent
\textbf{Sensor model.} The HES sense the superposition of both fields. Evaluating Eq.~\eqref{eq:bfield_superposition} at the set \(\mathbf{r}_{\mathrm{HES}}\) of the 50 sensor positions and stacking the results gives \(\mathbf{B}^{\mathrm{mover}}_t(\mathbf{r}_{\mathrm{HES}})\in\mathbb{R}^{150}\), so the observation is
\begin{equation}
  \mathbf{o}_t = c_{\mathrm{sens}}\left(\mathbf{B}^{\mathrm{mover}}_t(\mathbf{r}_{\mathrm{HES}}) + \mathbf{B}^{\mathrm{stator}}_t(\mathbf{i}_t)\right) + \boldsymbol{\eta}_t,
  \label{eq:sensor_model}
\end{equation}
with sensor scaling \(c_{\mathrm{sens}}\) and additive noise \(\boldsymbol{\eta}_t\).

\noindent
\textbf{Forces and Torques.} We compute the electromagnetic force on each
stator wire segment defined in Eq.~\eqref{eq:biot_savart_segment}, then
map it to the force and torque on the mover by Newton's third
law~\cite{Kim1998,Jansen2008,VanLierop2008}. The mover field at each
segment midpoint \mbox{\(\mathbf{r}_\ell=\tfrac{1}{2}(\mathbf{r}_\ell^-+\mathbf{r}_\ell^+)\)}
gives the force on segment \(\ell\)
\begin{equation}
  \mathbf{f}^{\mathrm{stator}}_{\ell,t}
  = I_j^s\,\Delta\mathbf{l}_\ell \times
    \mathbf{B}^{\mathrm{mover}}_t(\mathbf{r}_\ell),
  \label{eq:lorentz_segment}
\end{equation}
where \(I_j^s\) is the current of Eq.~\eqref{eq:dq_currents} assigned
to segment \(\ell\), and \(\Delta\mathbf{l}_\ell\) its directed
segment vector. By Newton's third law, the mover experiences the
opposite reaction, so at time step \(t\) the total mover force \(\mathbf{F}^{\mathrm{mover}}_t\) and torque \(\boldsymbol{\tau}^{\mathrm{mover}}_t\) are
\begin{align}
  \mathbf{F}^{\mathrm{mover}}_t
  &= -\sum_{\ell} \mathbf{f}^{\mathrm{stator}}_{\ell,t}
    + m_{\mathrm{mover}}\mathbf{g}
    + \mathbf{f}^{\mathrm{ext}}_t,
  \label{eq:force_kernel} \\
  \boldsymbol{\tau}^{\mathrm{mover}}_t
  &= -\sum_{\ell}
    \left(\mathbf{r}_\ell-\mathbf{s}^{\mathrm{CoM}}_t\right)
    \times
    \mathbf{f}^{\mathrm{stator}}_{\ell,t}
    + \boldsymbol{\tau}^{\mathrm{ext}}_t,
  \label{eq:torque_kernel}
\end{align}
where \(\mathbf{s}^{\mathrm{CoM}}_t\) is the center-of-mass position of the mover,
\(m_{\mathrm{mover}}\) is the mass of the mover, \(\mathbf{g}\) is gravity, and \(\mathbf{f}^{\mathrm{ext}}_t\),
\(\boldsymbol{\tau}^{\mathrm{ext}}_t\) optional external disturbances.
Since Eq.~\eqref{eq:lorentz_segment} is linear in current, the
electromagnetic wrench can be rephrased as \(\mathbf{A}_t\mathbf{i}_t\), with
the current-to-wrench matrix \(\mathbf{A}_t\in\mathbb{R}^{6\times72}\)
reassembled at the mover pose at each time step \(t\). The current command thus
enters only through \(\mathbf{A}_t\), so its gradient is exactly
\(\mathbf{A}_t\) and never backpropagates through the per-segment
field summation.
For efficiency, \(\mathbf{A}_t\) is held fixed over four control cycles, which we found to be the best trade-off between computation time and accuracy. On the training references, this speeds up training 3.5 times at relative force and torque errors of \SI{0.5}{\percent} and \SI{0.8}{\percent}.

\noindent
\textbf{Dynamics.} The forces and torques from Eq.~\eqref{eq:force_kernel}
and Eq.~\eqref{eq:torque_kernel} are applied over one control cycle
$\Delta t = \SI{250}{\micro\second}$ using semi-implicit Euler
integration~\cite{Lee2020}. Let \(m_{\mathrm{mover}}\) denote the
mover mass, $\mathbf{I}_{\mathrm{mover}}$ its constant body-frame
inertia tensor, and $\mathbf{I}_{\mathrm{w}} = \mathbf{R}_t
\mathbf{I}_{\mathrm{mover}} \mathbf{R}_t^\top$ the corresponding
world-frame inertia tensor. With diagonal linear and angular damping matrices $\mathbf{D}_v, \mathbf{D}_\omega \in \mathbb{R}^{3\times3}$, the mover state updates as
\begin{align}
  \mathbf{v}_{t+1} &= \mathbf{v}_t +
    \frac{\mathbf{F}^{\mathrm{mover}}_t - \mathbf{D}_v \mathbf{v}_t}{m_{\mathrm{mover}}}\,\Delta t, \\
  \mathbf{s}_{t+1} &= \mathbf{s}_t + \mathbf{v}_{t+1}\,\Delta t, \\
  \boldsymbol{\omega}_{t+1} &= \boldsymbol{\omega}_t +
    \mathbf{I}_{\mathrm{w}}^{-1}
    \big(
        \boldsymbol{\tau}^{\mathrm{mover}}_t - \boldsymbol{\tau}^{\mathrm{gyro}}_t - \mathbf{D}_\omega\,\boldsymbol{\omega}_t
    \big)\Delta t, \\
  \mathbf{R}_{t+1} &= e^{[\boldsymbol{\omega}_{t+1}]_\times \Delta t}\mathbf{R}_t,
\end{align}
where $\mathbf{v}_t,\boldsymbol{\omega}_t$ are the mover's linear and
angular velocity, $\boldsymbol{\tau}^{\mathrm{gyro}}_t =
\boldsymbol{\omega}_t \times (\mathbf{I}_{\mathrm{w}}\boldsymbol{\omega}_t)$
is the gyroscopic term from the Newton--Euler
equations~\cite{Holzinger2024}, and $[\boldsymbol{\omega}]_\times
\in \mathbb{R}^{3\times3}$ is the skew-symmetric matrix built from
$\boldsymbol{\omega}$, satisfying $[\boldsymbol{\omega}]_\times \mathbf{a}
= \boldsymbol{\omega} \times \mathbf{a}$ for any vector $\mathbf{a}$.
Updating the rotation matrix $\mathbf{R}$ via the exponential map is a
Lie group integration method that keeps the orientation on the
$SO(3)$ manifold without periodic re-normalization~\cite{Bruls2010}. Using the updated velocity $\mathbf{v}_{t+1}$ for the
position step improves numerical stability over explicit
Euler~\cite{Bruls2010,Lee2020}.
For contact handling, only mover--tile contact is considered, using the lowest point of the mover
\begin{equation}
z_{\min} = s_{z,t+1} - \lvert R_{31}\rvert h_x - \lvert R_{32}\rvert h_y - (\lvert R_{33}\rvert - R_{33})\,h_z,
\end{equation}
with \(R_{3j}\) the third row of \(\mathbf{R}_{t+1}\) and \(h_x,h_y,h_z\) half the mover's side lengths. On contact, the mover is lifted onto the tile and damped. Training integrates the translation at \(\mathbf{s}_t\), while rollouts (Sec.~\ref{sec:experiments}) use the center of mass and frictional contact.

\begin{figure}[t]
    \centering
    \includegraphics[width=\linewidth]{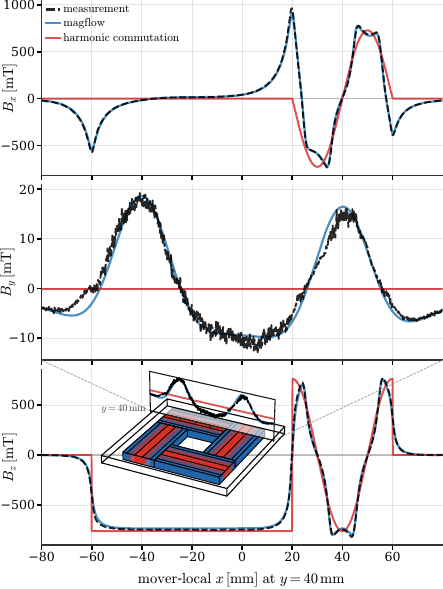}
    \vspace{-2.0em}
    \caption{Magnetic field components of the mover along a cut at mover-local \(y=\SI{40}{\milli\metre}\), \(z=\SI{-1}{\milli\metre}\), comparing measurements (black), the \magflow{} simulation (blue), and the harmonic commutation model (red). The first-harmonic model deviates systematically from the measurements, while \magflow{} matches them.}
    \vspace{-1.6em}
    \label{fig:halbach_first_harmonic_comparison}
\end{figure}

\subsection{Training of Neural Controller}

\begin{figure*}[t]
    \centering
    \includegraphics[width=1.0\textwidth]{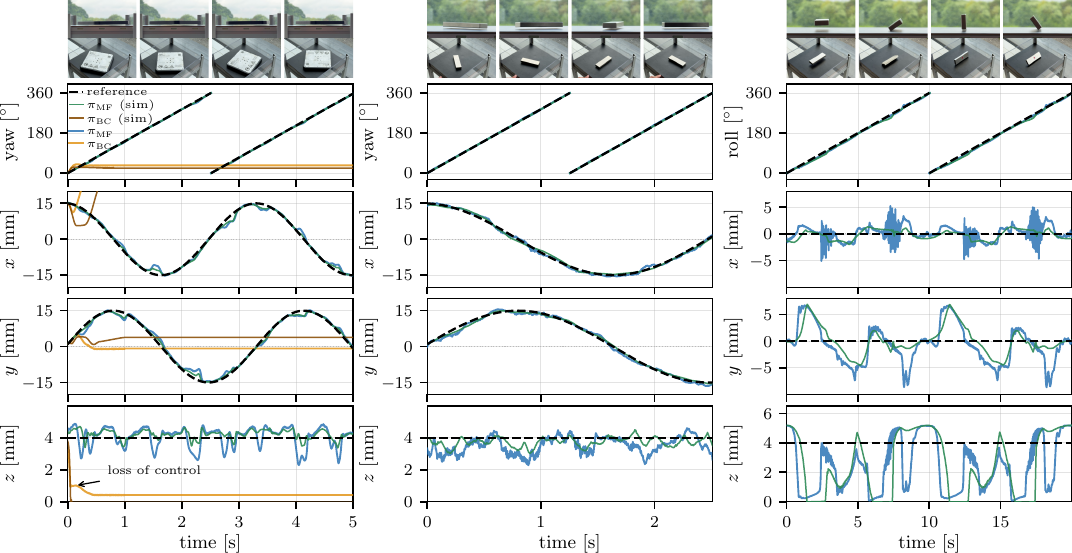}
    \vspace{-2.0em}
    \caption{Neural controllers trained end-to-end in \magflow{} (\piPHY{}, blue) and deployed zero-shot achieve novel capabilities. Closed-loop rollouts of the deployed controllers in \magflow{} on the same references are shown in green (\piPHY{}) and brown (\piNC{}). \textbf{(left)}~Full \ang{360} yaw rotation of the APM4330 mover away from the tile center. The behavior-cloned baseline \piNC{} (orange) directly loses control. \textbf{(middle)}~A neural controller retrained in \magflow{} for a plain bar magnet, a topology with no harmonic commutation model, levitates it stably and tracks a full \ang{360} yaw rotation away from the tile center. \textbf{(right)}~Continuous \ang{360} rotation of the bar magnet about a horizontal axis, to our knowledge not shown before for a single magnet on a planar MagLev system.}

    \vspace{-1.6em}
    \label{fig:experiments}
\end{figure*}

\noindent
\textbf{Inputs and reference.} Aligning to prior work~\cite{Hartmann2025},
the neural controller is a recurrent map from raw HES readings and
the reference pose to actuator commands. The reference is
encoded as \(\tilde{\mathbf{p}}_t\in\mathbb{R}^8\), stacking the position \((x,y,z)\), the mover surface normal
\((\mathbf{R}_{13},\mathbf{R}_{23},\mathbf{R}_{33})\), and the yaw angle \(\psi\) as \((\cos n\psi,\sin n\psi)\). Neglecting
manufacturing uncertainties, the APM4330 mover has \ang{90} yaw
symmetry, so \(n=4\) distinct yaw angles per full rotation are
physically indistinguishable configurations (\(n=2\) for the bar
magnet). Encoding yaw this way makes the reference invariant to this
symmetry, which we exploit to make learning more efficient. Therefore, full \ang{360} yaw rotation is
only realizable by continuously tracking through the repeated encoding
over time. With HES readings
\(\mathbf{o}_t\in\mathbb{R}^{150}\) and hidden state
\(\mathbf{h}_t\), the recurrent map is
\begin{equation}
  \mathbf{a}_t,\mathbf{h}_{t+1}
  = \piPHY(\mathbf{o}_t,\tilde{\mathbf{p}}_t,\mathbf{h}_t).
\end{equation}
\textbf{Action encoding.} Each stator unit is driven by
field-oriented control (Sec.~\ref{sec:system_hardware}) from a
triplet \((d_j,q_j,\phi_j)\). For each of the 24 stator units \(j\),
the neural controller \piPHY{} predicts a stationary-frame (Clarke)
current pair \((u_j,v_j)\), stacked into the 48-dimensional action
\(\mathbf{a}_t\). To convert this into the field-oriented control
required by the hardware interface, we compute
\(d_j=\lVert(u_j,v_j)\rVert_2\) and \(\phi_j=\operatorname{atan2}(v_j,u_j)\)
with \(q_j=0\). This
vector maps bijectively to the coil currents, unlike the
\((d,q,\sin\phi,\cos\phi)\) output of prior work~\cite{Hartmann2025}.
A controller trained from scratch has no canonical choice and can change between equivalent representations along a BPTT rollout, leading to gradient spikes. The new encoding avoids this.

\noindent
\textbf{Architecture and objective.} The architecture is a 4-layer gated recurrent unit (GRU) with 256 hidden units per layer, the same as the behavior-cloning proof of concept~\cite{Hartmann2025}, whose implementation optimized for Advanced Vector Extensions 2 (AVX2) has a worst-case inference latency of \SI{242}{\micro\second}, satisfying the \SI{250}{\micro\second} hard real-time constraint. Training uses truncated BPTT through the closed loop at the hardware control rate to match deployment. We use BPTT, since the smooth electromagnetic physics provide analytic gradients, whereas reinforcement learning would have to explore 48 continuous actions of an unstable system, where random actions quickly crash the mover, and assign credit over transients spanning hundreds of steps at \SI{4}{\kilo\hertz}. A controller trained at a lower rate would not match the fixed deployment cycle. Gradients flow over windows of \(K=500\) steps before each optimizer step (AdamW, learning rate \(5\!\times\!10^{-5}\), gradient clipping at 100), then the rollout continues from the carried hidden and physical state to the horizon \(H\).
A curriculum grows \(H\) from 100 steps by \(1.25\times\) once \(80\%\) of the worlds do not crash. Rollouts run in parallel across 16 randomized worlds, half tracking a hover with a discontinuous reference pose step and half a smooth random 6D trajectory over the whole tile.
In each world, a random external 6D wrench, a per-world constant bias plus an Ornstein--Uhlenbeck process, acts on the mover to make the learned neural controller \piPHY{} robust against external disturbances.
Training takes five days, as the full interaction of the 20 magnets of the APM4330 with \num{130368} coil segments (\(\|\Delta\mathbf{l}_\ell\|=\SI{3}{\milli\meter}\), Eq.~\eqref{eq:lorentz_segment}) is recomputed every four cycles. This fully utilizes an NVIDIA A40 at \SI{3}{\milli\second} per step including the backward pass.
The objective over the network parameters \(\theta\) sums per-step tracking, safety, and smoothness costs,
\begin{equation}
  \mathcal{L}(\theta)=\mathbb{E}\sum_{t=0}^{H}\big[\mathcal{L}_{\mathrm{track}}
    +\mathcal{L}_{\mathrm{safe}}+\mathcal{L}_{\mathrm{smooth}}\big],
\end{equation}
where \(\mathcal{L}_{\mathrm{track}}=\lambda_{\mathrm{pos}}\|\mathbf{s}_t-\mathbf{s}_t^\star\|_2^2+\lambda_{\mathrm{rot}}\,d_{SO(3)}^2(\mathbf{R}_t,\mathbf{R}_t^\star)\) with \(\lambda_{\mathrm{pos}}=\lambda_{\mathrm{rot}}=1\) penalizes position (\si{\milli\meter}) and geodesic orientation (\si{\degree}) error. The safety term \(\mathcal{L}_{\mathrm{safe}}\) combines four penalties. To avoid collisions, a sigmoid barrier \(\lambda_z\,\mathrm{sigmoid}\big((z_c-z_{\min})/T_z\big)\) with \(\lambda_z=10\) rises as the mover's lowest point \(z_{\min}\) approaches the threshold \(z_c=\SI{0.5}{\milli\meter}\), with temperature \(T_z=\SI{0.2}{\milli\meter}\). To protect the coils, \(\lambda_c\,\mathrm{relu}(d_j-I_{\max})^2\) with \(\lambda_c=500\) bounds the current of each unit to \(I_{\max}=\SI{33}{\ampere}\). It acts only above \(I_{\max}\), as a general current penalty would trade off against tracking, in particular against lift in \(z\).
Below \(I_{\max}\), the currents of coil units not covered by the mover remain unconstrained, as they barely affect the mover and thus no loss term depends on them. The locality penalty \(\sum_{j\in\mathcal{U}_t}(\lambda_1 d_j+\lambda_2 d_j^2)\) with \(\lambda_1=\lambda_2=5\times10^{-5}\) over the set \(\mathcal{U}_t\) of these uncovered units therefore drives their currents to zero. The power supply empirically permits \SI{140}{\watt} continuously and up to \SI{240}{\watt} for short periods, with the available limit \(P^{\lim}_t\) decreasing as a time-to-shutdown budget reported by the hardware. If the commanded power \(P_t\) exceeds it, all currents are derated by \(\gamma_t=\sqrt{P^{\lim}_t/P_t}\) as on the hardware, and \(\lambda_P\,\mathrm{relu}(1-\gamma_t)^2\) with \(\lambda_P=500\) penalizes this derating.
Finally, \(\mathcal{L}_{\mathrm{smooth}}=\lambda_{\Delta i}\|\Delta^2\mathbf{i}_t\|_2^2+\lambda_{\Delta w}\|\Delta^2(\mathbf{A}_t\mathbf{i}_t)\|_2^2\) with \(\lambda_{\Delta i}=1\) and \(\lambda_{\Delta w}=10\) penalizes the curvature of the commanded phase currents and of the realized wrench.

\noindent
\textbf{Domain randomization.}\label{subsec:domain_rand}
Real movers and tiles are subject to manufacturing tolerances, and
the real system's true parameters are unknown and vary from unit to
unit. To make the neural controller \piPHY{} robust against these
uncertainties, we follow related works~\cite{Tobin2017,Andrychowicz2019,Peng2018,GraefeCFB2026}
and apply domain randomization by sampling new physical parameters for
every world and rollout. We randomize the mover (magnet remanence, per-magnet position
and orientation, mass, inertia, and center of mass), the sensing
(per-HES position and measurement noise), and the actuation (per-coil
current gain and the current lag time), together with the per-rollout
external wrench disturbance. The exact ranges are in the released
code.

\section{Experimental Evaluation}
\label{sec:experiments}

\begin{table}[t]
\centering
\caption{RMS tracking error on a random trajectory measured by motion capture. Sim rows are closed-loop \magflow{} rollouts.}
\label{tab:randtraj}
\vspace{-0.8em}
\setlength{\tabcolsep}{6pt}
\begin{tabular}{l ccc ccc}
 & \multicolumn{3}{c}{Position [mm]} & \multicolumn{3}{c}{Orientation [$^\circ$]} \\
\cmidrule(lr){2-4}\cmidrule(lr){5-7}
Controller & $x$ & $y$ & $z$ & roll & pitch & yaw \\
\midrule
$\pi_{\mathrm{PC}}$       & \textbf{0.09} & \textbf{0.07} & \textbf{0.06} & \textbf{0.10} & \textbf{0.07} & \textbf{0.05} \\
$\pi_{\mathrm{BC}}$       & 0.15 & 0.14 & 0.09 & 0.12 & 0.11 & 0.10 \\
$\pi_{\mathrm{MF}}$       & 0.34 & 0.30 & 0.14 & 0.24 & 0.25 & 0.26 \\
\midrule
$\pi_{\mathrm{BC}}$ (sim) & 0.16 & 0.18 & 0.08 & 0.10 & 0.10 & 0.24 \\
$\pi_{\mathrm{MF}}$ (sim) & 0.31 & 0.24 & 0.15 & 0.19 & 0.21 & 0.22 \\
\bottomrule
\end{tabular}
\vspace{-1.6em}
\end{table}

\noindent
Our evaluation consists of three parts. We first compare the physics behind the two control paradigms. Engineered controllers compute currents from the reduced first-harmonic model, while \magflow{} trains the neural controller \piPHY{} using a magnetic field simulation.
We then deploy the three neural controllers trained in \magflow{} on the real hardware, show qualitatively new motions never before demonstrated on this MagLev system, and evaluate their accuracy within the engineered range and their robustness to payload.

\subsection{First-Harmonic Model vs.\ \magflow{}}
\label{subsec:field_accuracy}

\noindent
We compare the first-harmonic model of Eq.~\eqref{eq:first_harmonic_fields} and the \magflow{} field of Eq.~\eqref{eq:bfield_superposition} against measurements of the real APM4330 mover. Fig.~\ref{fig:halbach_first_harmonic_comparison} shows the magnetic field along a cut below the mover. The first-harmonic model deviates systematically for the reasons discussed in Sec.~\ref{subsec:model_mismatch}. \magflow{} matches the measurements closely along the entire cut. The neural controller is therefore trained against a field model that reproduces the hardware, instead of the approximated model that hand-crafted controllers use.

\subsection{Zero-Shot Capabilities Beyond Engineered Control}
\label{subsec:capabilities}

\noindent
We train three neural controllers \piPHY{} exclusively in \magflow{}, each for five days on a single NVIDIA A40 GPU, and deploy them zero-shot on the real system at \SI{4}{\kilo\hertz}. The first controls the APM4330 with full \ang{360} yaw over the whole tile, the second the bar magnet with full \ang{360} yaw, and the third the bar magnet with full \ang{360} yaw and roll. Aligning to prior work~\cite{Hartmann2025}, reported poses are HES-based estimates, except for Table~\ref{tab:randtraj}, which uses VICON-based motion capture. Fig.~\ref{fig:experiments} shows three experiments, each demonstrating a capability that no previous controller of this platform has shown, together with \magflow{} closed-loop simulations of the deployed controllers on the same references.

\noindent
\textbf{Yaw beyond engineered limits.} We command a full \ang{360} yaw rotation of the APM4330 mover while tracking a circle in $xy$ with \SI{15}{\milli\meter} radius. \piPHY{} tracks the full rotation without loss of control. The behavior-cloned controller \piNC{}~\cite{Hartmann2025} inherits the workspace of its teacher and immediately loses control.

\noindent
\textbf{Bar magnet topology.} We retarget training to a plain bar magnet (\(100\times30\times10\,\si{\milli\metre}\), neodymium N48) by replacing the mover model in \magflow{} and retraining. For this topology the co-design assumptions do not hold and no engineered controller exists (Sec.~\ref{subsec:model_mismatch}). The retrained controller levitates the bar magnet and tracks a full \ang{360} yaw rotation while tracking a circle in $xy$ with \SI{15}{\milli\meter} radius. Unlike a disc magnet, the bar magnet is controlled in 6D, including yaw.

\begin{figure}[t]
    \centering
    \includegraphics[width=\linewidth]{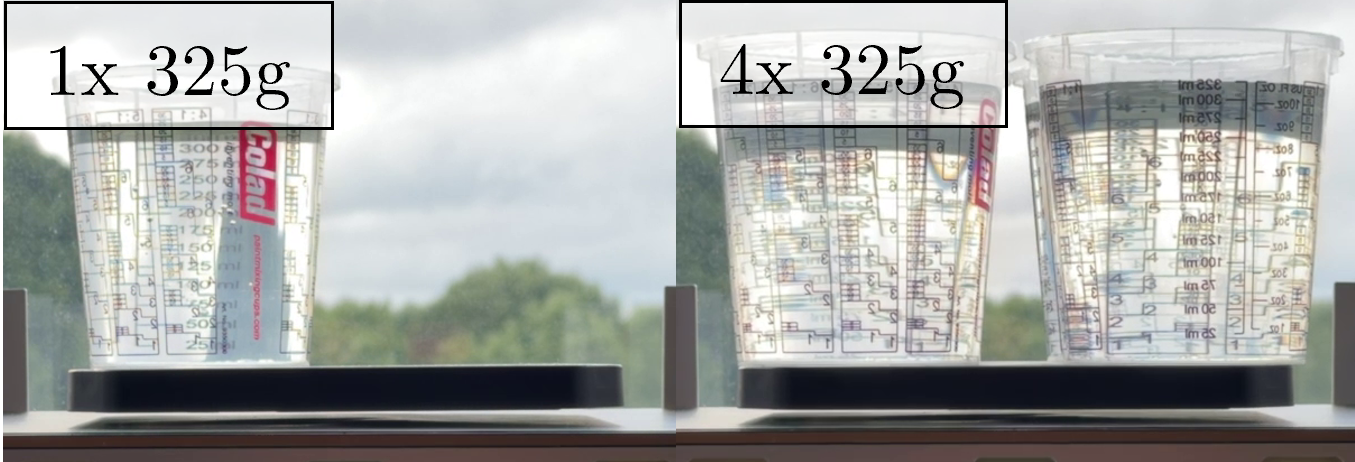}
    \vspace{-2.0em}
    \caption{The neural controller \piPHY{} hovers the APM4330 mover stably under a tested payload of \SI{1.3}{\kilo\gram}.}
    \vspace{-1.6em}
    \label{fig:weights}
\end{figure}

\noindent
\textbf{Rotation about a horizontal axis.} The neural controller further rotates the levitating bar magnet \ang{360} about a horizontal axis. During this rotation, the bar magnet occasionally touches the tile. Unlimited rotation was previously only shown for a co-designed body of six magnets with optical tracking~\cite{MiyasakaBerkelman2014}. To our knowledge, this is the first continuous \ang{360} rotation of a single magnet about a horizontal axis on a planar MagLev system, with pose sensing only by the embedded HES.

\noindent
\textbf{Accuracy within the engineered range.} Table~\ref{tab:randtraj} compares all controllers on a smooth random \SI{60}{\second} trajectory, measured with a Vicon motion capture system. It moves continuously within the workspace of \piPC{} and \piNC{} at a median speed of \SI{33}{\milli\meter\per\second}, whereas \piPHY{} is trained for full \ang{360} yaw. \piPC{} is the proprietary controller, engineered and tuned over decades, and \piNC{} imitates it~\cite{Hartmann2025}. \piPHY{} is less precise than both, but learns purely from simulated physics without any engineered module or teacher. Its closed-loop rollouts in \magflow{} match the real position errors within \SI{0.06}{\milli\meter}, a gap far smaller than the gap to \piPC{}. Precision is therefore limited by the neural controller rather than by \magflow{}.

\subsection{Robustness to Payload}
\label{subsec:robustness}

\noindent
The random 6D wrench disturbance \((\mathbf{f}^{\mathrm{ext}}_t,\boldsymbol{\tau}^{\mathrm{ext}}_t)\) of Eqs.~\eqref{eq:force_kernel} and~\eqref{eq:torque_kernel} and the randomized mover mass (Sec.~\ref{subsec:domain_rand}) make \piPHY{} robust against external disturbances. Fig.~\ref{fig:weights} shows that this robustness extends to real payloads, with the APM4330 mover hovering stably under a tested payload of \SI{1.3}{\kilo\gram}.

\section{Conclusion and Future Work}
\label{sec:conclusion}

\noindent
\magflow{} moves the electromagnetic computation of 6D MagLev control out of the real-time loop and into offline training. The resulting neural controllers transfer zero-shot to industrial hardware and achieve capabilities that no prior controller of the utilized MagLev system has shown. Future work can extend \magflow{} in several directions. Scaling to multi-tiles can exploit the symmetries of sensing and actuation across tiles, but requires handling movers that cross tile boundaries. Multi-movers are even more demanding, as nearby movers potentially share coils and Hall sensors. Closing the precision gap to engineered control, which our results attribute to the neural controller rather than to \magflow{}, calls for larger or more expressive neural controllers, enabled by faster inference hardware within the \SI{250}{\micro\second} cycle. However, quantifying such improvements at the micrometer scale may require precise external measurements like laser interferometry. \magflow{} also approximates the physics. It models coils as thin wires and magnets as ideal, uniformly magnetized cuboids, neglects ferromagnetic material and demagnetization, and treats the mover as a rigid body. Calibrating sensors and magnets with real data, or learning a residual correction of the simulation~\cite{PanLearningFly2026}, could further reduce the gap between simulation and reality. As a GPU-parallel simulator, \magflow{} equally supports other learning schemes such as reinforcement learning, which could be compared with or combined with BPTT. Finally, predicting action chunks or refining the simulation frequency as a curriculum could reduce inference and training cost.

\bibliographystyle{IEEEtran}
\bibliography{IEEEabrv,references}

\end{document}